\documentclass[11pt,a4paper]{article}
\usepackage[hyperref]{acl2019}
\usepackage{times}
\usepackage{latexsym}
\usepackage{amssymb}
\usepackage{amsmath}
\usepackage{tabularx}
\usepackage{booktabs}
\usepackage{graphicx}
\usepackage{xspace}
\usepackage{url}
\usepackage{adjustbox}

\aclfinalcopy 

\newcommand\norm[1]{\left\lVert#1\right\rVert}
\newcommand{\TPRgap}{TPR\textsubscript{gap}}
\newcommand{\TNRgap}{TNR\textsubscript{gap}}

\title{Debiasing Embeddings for Reduced Gender Bias in Text Classification}

\author{Flavien Prost\thanks{\, Equal contribution.} \\
  \texttt{fprost@google.com} \\\And
  Nithum Thain\footnotemark[1] \\
  \texttt{nthain@google.com} \\\And
  Tolga Bolukbasi \\
  \texttt{tolgab@google.com} \\}

\date{}

\begin{document}
\maketitle
\begin{abstract}

\cite{bolukbasi2016man} demonstrated that pretrained word embeddings can inherit gender bias from the data they were trained on. We investigate how this bias affects downstream classification tasks, using the case study of occupation classification \cite{de2019bias}. We show that traditional techniques for debiasing embeddings can actually worsen the bias of the downstream classifier by providing a less noisy channel for communicating gender information. With a relatively minor adjustment, however, we show how these same techniques can be used to simultaneously reduce bias and maintain high classification accuracy.

\end{abstract}

\section{Introduction}
A trend in the construction of deep learning models for natural language processing tasks is the use of pre-trained embeddings at the input layer \cite{mikolov2013efficient, pennington2014glove, bojanowski2017enriching}. These embeddings are usually learned by solving a language modeling task on a large unsupervised corpus, allowing downstream models to leverage the semantic and syntactic relationships learned from this corpus. One issue with using such embeddings, however, is that the model might inherit unintended biases from this corpus. In \cite{bolukbasi2016man}, the authors highlight some gender bias at the embedding layer through analogy and occupational stereotyping tasks, but do not investigate how these biases affect modeling on downstream tasks. It has been argued \cite{gonen2019lipstick} that such debiasing approaches only mask the bias in embeddings and that bias remains in a form that downstream algorithms can still pick up.

This paper investigate the impact of gender bias in these pre-trained word embeddings on downstream modeling tasks. We build deep neural network classifiers to perform occupation classification on the recently released ``Bias in Bios'' dataset \cite{de2019bias} using a variety of different debiasing techniques for these embeddings introduced in \cite{bolukbasi2016man} and comparing them to the scrubbing of gender indicators. The main contributions of this paper are:
\begin{itemize}
    \item Comparing the efficacy of embedding based debiasing techniques to manual word scrubbing techniques on both overall model performance and fairness.
    \item Demonstrating that standard debiasing approaches like those introduced in \cite{bolukbasi2016man} actually worsen the bias of downstream tasks by providing a denoised channel for communicating demographic information.
    \item Highlight that a simple modification of this debiasing technique which aims to completely remove gender information can simultaneously improve fairness criteria and maintain a high level of task accuracy.
\end{itemize}



\section{Classification Task}

This work utilizes the BiosBias dataset introduced in \cite{de2019bias}. This dataset consists of biographies identified within the Common Crawl. 397,340 biographies were extracted from sixteen crawls from 2014 to 2018. Biography lengths ranged from eighteen to 194 tokens and were labelled with one of twenty-eight different occupations and a binary gender (see Table \ref{tab:dataset} for a more detailed breakdown of statistics). The goal of the task is to correctly classify the subject's occupation from their biography. Each comment is assigned to our train, dev, and test split with probability 0.7, 0.15, and 0.15 respectively. 

In \cite{de2019bias}, this task is used to explore the level of bias present in three different types of models: a bag of words logistic regression, a word embedding logistic regression, and a bi-directional recurrent neural network with attention. The models are trained with two different tokenization strategies, i.e. with and without scrubbing gender indicators like pronouns. We will use the two variants of the deep neural network model as a baseline in this work.

\begin{table}[t]
    \centering
    \begin{tabular}{lrr}
    \toprule
    \textbf{Occupation}    & \textbf{Female Bios}  & \textbf{Male Bios}\\
    \midrule
        accountant &   3579 &   2085 \\
         architect &   7747 &   2409 \\
          attorney &  20182 &  12531 \\
      chiropractor &   1973 &    705 \\
          comedian &   2223 &    594 \\
          composer &   4700 &    921 \\
           dentist &   9573 &   5240 \\
         dietitian &    289 &   3696 \\
                dj &   1279 &    211 \\
         filmmaker &   4712 &   2314 \\
 interior designer &    282 &   1185 \\
        journalist &  10110 &   9896 \\
             model &   1295 &   6244 \\
             nurse &   1738 &  17263 \\
           painter &   4210 &   3550 \\
         paralegal &    268 &   1503 \\
            pastor &   1926 &    609 \\
  personal trainer &    782 &    656 \\
      photographer &  15669 &   8713 \\
         physician &  20805 &  20298 \\
              poet &   3587 &   3448 \\
         professor &  65049 &  53438 \\
      psychologist &   7001 &  11476 \\
            rapper &   1274 &    136 \\
 software engineer &   5837 &   1096 \\
           surgeon &  11637 &   2023 \\
           teacher &   6460 &   9813 \\
      yoga teacher &    259 &   1408 \\
    \bottomrule
    \end{tabular}
    \caption{Dataset Statistics}
    \label{tab:dataset}
\end{table}





% ...

\section{Debiasing Methodology}
\label{sec:debias}

\subsection{Debiasing Word Embeddings} \label{debiasingsection}

Our DNN models use 100 dimensional normalized GloVe embeddings \cite{pennington2014glove} at the input layer. \cite{bolukbasi2016man} showed through analogy and occupational stereotyping tasks that such embeddings contain instances of direct and indirect bias. 
They also provide a technique that can be used to remove this bias as measured by this task. 
In this section, we review the specifics of this technique.

The first step to produce debiased word embeddings from our input GloVe embeddings $\{ \vec{w} \in \mathbb{R}^d \}$ is to define a collection of word-pairs $D_1, ..., D_n$ which can be used to identify the gender subspace. For this work, we use the same input word pairs as \cite{bolukbasi2016man}. The $k$-dimensional gender subspace $B$ is then defined to be the first $k$ rows of the singular value decomposition of 

$$\frac{1}{2} \sum_{i=1}^n \sum_{\vec{w} \in D_i} (\vec{w} - \mu_i)^T (\vec{w} - \mu_i)$$

\noindent where $\mu_i := \sum_{\vec{w} \in D_i} \vec{w}/2$. For our experiments, we set $k = 1$.

The next step is to modify the embedding of a set of ``neutral'' (i.e. non-gendered) words $N$ by projecting them orthogonally to the gender subspace. If we let $\vec{w}_B$ denote the projection of a word embedding $\vec{w}$ orthogonally to the gender subspace $B$ then this would be equivalent to, for all neutral word vectors $\vec{w} \in N$, changing their embedding to:

$$\vec{w} := \vec{w}_B/\norm{\vec{w}_B}.$$

The final step in the algorithm is to define a collection of equality sets $E_1, ..., E_m$ of words which we believe should differ only in the gender component. For our purposes we use all the word pairs used in \cite{bolukbasi2016man} as well as the sets of words that are scrubbed in \cite{de2019bias}. For each $E_i$ we equalize by taking the mean $\mu = \sum_{\vec{w} \in E_i} \vec{w}/|E_i|$ and projecting that orthogonally to the gender subspace to obtain $\mu_B$. The new embeddings for each word in the equalize set $\vec{w} \in E_i$ can then be set to

$$\mu_B + \sqrt{1 - \norm{\mu_B}^2} \frac{\vec{w_B}}{\norm{\vec{w_B}}}.$$

To compute these debiased embeddings, we build on the github library\footnote{https://github.com/tolga-b/debiaswe} provided by the authors of \cite{bolukbasi2016man}.

\subsection{Strong Debiasing}

In the original work, \cite{bolukbasi2016man} differentiate between neutral words in the set $N$ and gender specific words, removing the gender subspace component of the former while preserving it for the latter. While this appears to be a good strategy for maintaining the maximum semantic information in the embeddings while removing as much biased gender information as possible, we show in Sections \ref{sec:experiments} and \ref{sec:analysis} that, by providing a clear channel to communicate gender information, this technique can make the gender bias worse in downstream modeling tasks. 

To mitigate this effect, we study \textit{strongly debiased embeddings}, a variant of the algorithm in the previous section where we simply set $N$ to be all of the words in our vocabulary. In this case, all words including those typically associated with gender (e.g. he, she, mr., mrs.) are projected orthogonally to the gender subspace. This seeks to remove entirely the gender information from the corpus while still maintaining the remaining semantic information about these words. It should be noted that for words in our equalize sets, i.e. those that differ only by gender, this results in all the words within one set being embedded to the same vector. As we will see in Section \ref{sec:experiments}, this results in an improved performance over techniques like scrubbing which remove this semantic information entirely and disrupt the language model of the input sentence. In Section \ref{subsec:ablation}, we also perform ablation studies to show how important each of the steps in the algorithm is to achieving high accuracy with low bias.


\section{Evaluation Metrics}

We will evaluate our models on the dimensions of overall performance and fairness. For the overall performance of these models, we will use the standard accuracy metric of multi-class classification. There are a number of metrics and criteria that offer different interpretations of model fairness \cite{dixon2018measuring, narayanan2018translation, friedler2019comparative, beutel2019putting}. In this work, we use the method introduced by \cite{hardt2016equality} as Equality of Opportunity. 

If your data has binary labels $Y$ and some demographic variable $A$, in our case whether the biography is about a female, then Equality of Opportunity is defined as

\begin{equation*}
\resizebox{1.0\hsize}{!}{$\Pr \{\hat{Y} = 1 | Y = 1, A = 1 \} = \Pr \{\hat{Y} = 1 | Y = 1, A = 0 \}$.}
\end{equation*}

i.e. the true positive rate of the model should be independent of the demographic variable conditioned on the true label.

In order to measure deviation from this ideal criteria we follow a number of other authors \cite{gargcounterfactual} and define the True Positive Rate Gap (\TPRgap) to be:

\begin{equation*}
\resizebox{1.0\hsize}{!}{$|\Pr \{\hat{Y} = 1 | Y = 1, A = 1 \} - \Pr \{\hat{Y} = 1 | Y = 1, A = 0 \}|$.}
\end{equation*}

Since in our context, we are dealing with a multi-class classifier, we will measure the \TPRgap \xspace for each class as a separate binary decision and will aggregate by averaging over all occupations. 

We can analagously define the \TNRgap \xspace with the True Negative Rates taking the place of True Positive Rates in the above discussion. We will also report the average \TNRgap \xspace across occupations.


\section{Experiments}
\label{sec:experiments}

To understand how the embedding layer affects our deep learning classifiers, we will train classifiers with a variety of embeddings. As baselines, we will use the GloVe embeddings with and without the gender indicator scrubbing described in \cite{de2019bias}. Additionally, we train a classifier on GloVe embeddings debiased using both techniques discussed in Section \ref{sec:debias}. These embeddings are fixed (rather than trainable) parameters of our network. For each of these models, we evaluate their classification performance (accuracy) alongside their overall fairness (\TPRgap).


\subsection{Model architecture}
Our architecture follows the DNN approach used in \cite{de2019bias}. 
After tokenization and embedding, we encode the input sentence with a bidirectional Recurrent Neural Network with GRU cells and extract the sentence representation by applying an attention layer over the bi-RNN outputs. After a dense layer with Relu activation, we compute a logit for each class via a linear layer.
We use the softmax cross-entropy to compute the loss.

All hyper parameters were tuned for the standard GloVe model and the optimal values were used for the subsequent runs.

\subsection{Scrubbing explicit gender indicators}

As a baseline with which to compare embedding based debiasing, we implement the scrubbing technique described in \cite{de2019bias}, which consists of preprocessing the text by removing explicit gender indicators.
To provide a fair comparison, we scrub explicit gender indicators that combine all the equalizing pairs used in (Bolukbasi et al., 2016) as well as the sets of words that are scrubbed in (De-Arteaga et al.,2019).

\subsection{Results}

Our results are displayed in Table \ref{tab:modelmetrics} which records the values of accuracy, {\TPRgap} and {\TNRgap} for each model. 
We focus on the {\TPRgap} as our primary fairness metric but the results still hold if the {\TNRgap} is used instead.

\begin{table}[t]
    \centering
    \begin{tabular}{lrrr}
    \toprule
    \textbf{Embedding}    & \textbf{Acc.}  & \textbf{\TPRgap} &  \textbf{\TNRgap}\\
    \midrule
     GloVe & 0.818 & 0.091 & 0.0031\\
     Scrubbed & 0.804 & 0.070 & 0.0024\\ 
     Debiased & 0.807 & 0.119 & 0.0037\\
     Strongly debiased & 0.817 & 0.069 & 0.0023\\
    \bottomrule
    \end{tabular}
    \caption{Model metrics}
    \label{tab:modelmetrics}
\end{table}

As we can see, the strongly debiased model performs best overall. It reduces the bias (-.022 {\TPRgap}) slightly more than the scrubbed technique (-.021 {\TPRgap}). However, it has a much smaller cost to classification accuracy than scrubbing (-0.1\% vs -1.4\%) as the embeddings still retain most of the semantic and syntactic information about the words in the comment.

Our results also indicate that using debiased embeddings has a counter-productive effect on bias as they significantly increase the {\TPRgap}. We explore this further in the next section.


\section{Analysis}
\label{sec:analysis}

In this section, we provide some analysis to explain our experimental results. 
First, we look at the sentence representation of each model and measure how much gender information it contains. We confirm empirically that both strongly debiasing and scrubbing techniques reduce the gender information that is used by the model.
Secondly, we explain why using debiased embeddings does not reduce the amount of bias in the classification task and highlight how the algorithm for strongly debiasing addresses this issue.
Finally, we explore the relative importance of the two components of the debiasing algorithm: projection and equalization.

\subsection{Connecting bias to gender information}
\label{subsec:genderinfo}

The debiasing techniques studied in this paper all modify the way the gender information is fed to the model. The scrubbed technique masks the explicit gender indicators, the debiased algorithm reduces the indirect gender bias in the embeddings and the strongly debiased approach aims at removing the gender component in them entirely.

We investigate how each of these embeddings affect the amount of gender information available to the model for occupation classification. To that end, after each model is trained on occupation labels, we train a separate logistic classifier that takes as input the sentence representation of the previous model (i.e. the last layer before the logits) and predicts the gender of the subject of the biography. We keep the remaining model layers frozen so that we do not change the representation of the sentence.
The accuracy of these gender classifiers (reported in Table \ref{tab:genderaccuracy}) provide a measure of the amount of gender information contained in each of our models. 

\begin{table}[t]
    \centering
    \begin{tabular}{lr}
    \toprule
    \textbf{Embedding}    & \textbf{Accuracy}\\
    \midrule
     GloVe & 0.86 \\
     Scrubbed & 0.68 \\
     Debiased & 0.88 \\
     Strong Debiased & 0.66 \\
    \bottomrule
    \end{tabular}
    \caption{Gender classifier accuracy}
    \label{tab:genderaccuracy}
\end{table}

We see that the representation of the baseline GloVe model keeps a significant amount of gender information, allowing for a gender classification accuracy of 0.86. With strongly debiased embeddings, the model representation contains much less gender information than the GloVe model and slightly less than scrubbing out the gender indicators. This suggests that both scrubbing and strong debiasing are effective at reducing the amount of gender information the overall model is able to capture, which explains the fairer occupational classifications.
Debiased embeddings, on the other hand, slightly increase the gender information that the model can learn. In the next section, we investigate this phenomenon further.

\subsection{Impact of debiasing on the gender component}

\begin{figure}
  \centering
  \includegraphics[width=\linewidth]{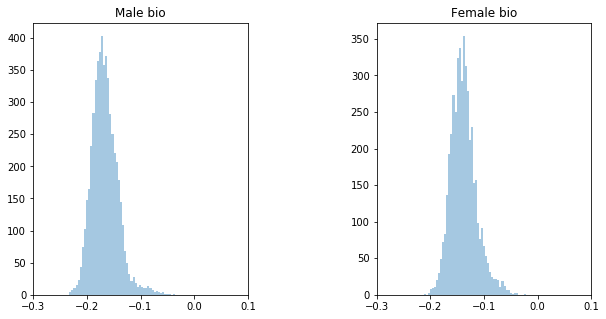}
  \caption{Gender component of a biography}
  \label{biosgendercomponent_all}
\end{figure}

The above experiments show that debiased embeddings cause models to act ``less fairly" than standard Glove embeddings (20\% higher \TPRgap) by allowing them to better represent the gender of the subject of the biography. 
We hypothesize that this is due to an undesirable side-effect of the debiasing algorithm introduced in Section \ref{debiasingsection}: it clarifies information coming from gender specific words by removing the noise from coming neutral words and therefore makes it easier for the model to communicate gender features.

To validate this hypothesis, we run the following analysis. Beginning with the standard GloVe embeddings, we define the gender component of a word as its projection on the gender direction and the gender component of a biography to be the average gender component of all the words it contains.

Figure \ref{biosgendercomponent_all} is a histogram of the gender component of the biographies in our data. A negative value means that the gender component of the biography is more male than female. We observe that, surprisingly, all biographies have a negative gender component \footnote{By comparing to other large datasets, we've established that this is an idiosyncrasy of the BiosBias dataset, which overall contains words with a more male gender component.}.

As expected, male biographies have a slightly more negative gender component on average than female biographies (-.166 vs -.138), which indicates that they include words that are more associated with male concepts. However, both distributions have a high variance and they are therefore not clearly separable. This plot indicates that, with standard GloVe embeddings, the gender component of a biography is only a weak signal for its gender.

\begin{figure}
  \centering
  \includegraphics[width=\linewidth]{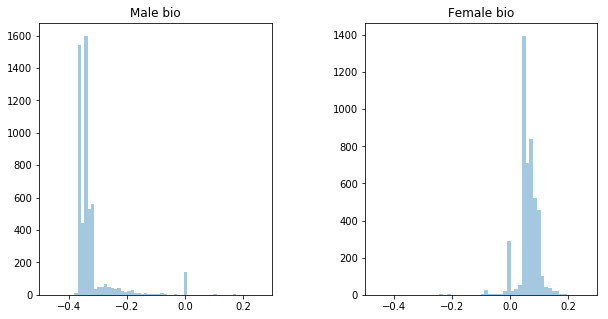}
  \includegraphics[width=\linewidth]{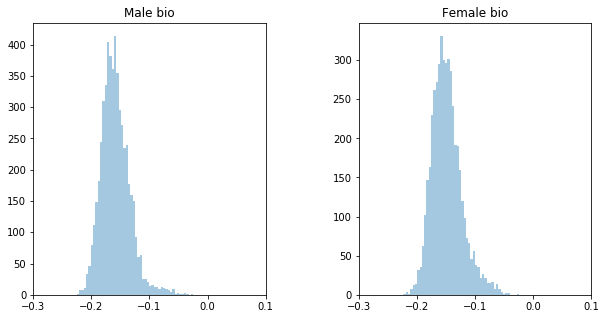}
  \caption{Gender component of a biography based on gender specific words (top) and based on neutral words (bottom)}
  \label{biosgendercomponent_split}
\end{figure}

Figure \ref{biosgendercomponent_split} is a histogram of the gender components of biographies which distinguishes the gender-specific words (top) and the neutral words (bottom).
We see that the gender component based on gender specific words gives a clear separation between male and female biographies, with a threshold at 0, whereas the distribution for neutral words (bottom) has more noise and does not clearly indicate the gender of the biography. Interestingly, for neutral words, the average gender component for male biographies is lower than for female ones (-.159 vs -.148), which indicates some indirect bias.

As about 95\% of the words of a biography are neutral, its gender component is mostly driven by this set of words and not by the gender-specific ones.
This explains why the plot in Figure \ref{biosgendercomponent_all} has a noisy distribution without any clear gender separation.
In other words, when using regular Glove embeddings, the neutral words are actually masking the clearer signal coming from gender specific words.

This analysis provides some explanation for the counter-intuitive impact of debiased embeddings. By construction, they remove the gender component of neutral words and leave unchanged that of gender specific words. While this is desirable for analogy tasks, using these embeddings in a text classifier actually allows the model to easily identify the gender of a biography and potentially learn a direct relationship between gender and occupation. On the opposite end, strongly debiased embeddings remove the entire signal and make it harder for the classifier to learn any such relationship.

\subsection{Ablation analysis of debiasing}
\label{subsec:ablation}

As mentioned in Section \ref{debiasingsection}, the algorithm for strongly debiasing includes two successive steps. First, we project all words orthogonally to the gender subspace. Then we equalize the non-gender part of a predefined list of pairs. 
We conducted an ablation study to analyze the impact of each step separately. More precisely, we train one model \textit{project only} where we project all the words orthogonally to the gender direction and another one  \textit{equalize only} where we equalize all pairs - which is equivalent to replacing each element of a pair by the mean vector.

\begin{table}[t]
    \centering
    \begin{tabular}{lrrr}
    \toprule
    \textbf{Embeddings} & \textbf{Acc.} & \textbf{\TPRgap} & \textbf{\TNRgap}\\
    \midrule
     GloVe & 0.818 & 0.091 & 0.0031 \\
     Strongly debias & 0.817 & 0.069 & 0.0023\\
     Project only & 0.815 & 0.103 & 0.0032\\
     Equalize only & 0.817 & 0.080 & 0.0029\\
    \bottomrule
    \end{tabular}
    \caption{Ablation study: Metrics for projection and equalization step}
    \label{tab:ablationtable}
\end{table}

Results are displayed in Table \ref{tab:ablationtable}. We observe that the equalization step has the strongest impact in bias reduction, while the projection is inefficient when used separately. We hypothesize that the projection is not able to correctly handle the explicit gender indicator words and therefore leaves too much direct bias. However both combined as in the strong debias technique provide the best results.

\section{Conclusion}
In this paper, we investigate how debiased embeddings affect both performance and fairness metrics. Our experiments reveal that debiased embeddings can actually worsen a text classifier's fairness, whereas strongly debiased embeddings can reduce gender information and improve fairness while maintaining good classification performance.
As these embeddings provide a simple tool that can be injected as is within model architectures, they do not result in much additional burden for ML practitioners (e.g. model tweaks, labelled data).
In the future, we would like to confirm that this approach generalizes to a variety of datasets and to other identity groups.



\bibliography{debias_bib}
\bibliographystyle{acl_natbib}

\end{document}